# An Attention-Based Multi-Context Convolutional Encoder-Decoder Neural Network for Work Zone Traffic Impact Prediction

Qinhua Jiang, Xishun Liao*, Yaofa Gong, and Jiaqi Ma *Senior Member, IEEE*

*Abstract*— Work zone is one of the major causes of non-recurrent traffic congestion and road incidents. Despite the significance of its impact, studies on predicting the traffic impact of work zones remain scarce. In this paper, we propose a data integration pipeline that enhances the utilization of work zone and traffic data from diversified platforms, and introduce a novel deep learning model to predict the traffic speed and incident likelihood during planned work zone events. The proposed model transforms traffic patterns into 2D space-time images for both model input and output and employs an attention-based multi-context convolutional encoder-decoder architecture to capture the spatial-temporal dependencies between work zone events and traffic variations. Trained and validated on four years of archived work zone traffic data from Maryland, USA, the model demonstrates superior performance over baseline models in predicting traffic speed, incident likelihood, and inferred traffic attributes such as queue length and congestion timings (i.e., start time and duration). Specifically, the proposed model outperforms the baseline models by reducing the prediction error of traffic speed by 5% to 34%, queue length by 11% to 29%, congestion timing by 6% to 17%, and increasing the accuracy of incident predictions by 5% to 7%. Consequently, this model offers substantial promise for enhancing the planning and traffic management of work zones.

## I. Introduction

The rising load on road infrastructures driven by population growth has resulted in an increased demand for road maintenance and reconstruction activities [1]. These work zone events often involve lane closures that will lead to more traffic crashes and delays caused by reduced road capacity. Unlike the usual congestion seen during peak traffic hours, work zone activities usually create non-recurring, unexpected travel delays. According to the Federal Highway Administration [2], work zones account for nearly 24% of non-recurring traffic congestion. Additionally, work zone events significantly endanger the safety of both travelers and workers; for instance, in 2022, traffic accidents in work zones resulted in 891 fatalities [2].

To address safety and mobility requirements during highway maintenance and construction, and to align with the expectations of the travelers, it is important for traffic management and work zone planning agencies to have an accurate estimation of how work zone events would impact the traffic. Modeling and predicting work zone impacts can enhance agency's decision-making as well as their overall understanding of the factors affecting work zone decisions [3].

Research on predicting the impact of work zones on traffic is very limited. Over the past few decades, related studies can be broadly divided into two categories: simulation or parametric-based approaches, and non-parametric, data-driven approaches.

In the field of simulation-based studies, Ping and Zhu [4] estimated the changes in traffic capacity under various work zones using CORSIM. Chatterjee et al. [5] considered drivers' behavior into simulation and developed a work zone traffic flow estimation model in VISSIM. Wen [6] developed a work zone traffic simulation model dedicated for connected traffic conditions. These simulation-based models usually only consider a few work zone factors and network configurations, thus are mostly unable to predict traffic conditions under unseen work zones with complex spatial-temporal patterns.

As the availability of data expands, facilitated by development in sensors and data collection techniques, the focus of research is increasingly turning towards data-driven methods, even though these data are not yet fully integrated. On the data-driven side, Adeli and Jiang [7] created a neuro-fuzzy model to estimate the traffic flow impacted by work zones. The results demonstrated the model's superiority over empirical approaches. Karim and Adeli [8] proposed an adaptive neural network model to predict the traffic impact including capacity, queue length, and delay during work zones. Hou et al. [9] developed four machine learning based work zone traffic prediction models: random forest, baseline predictor, regression tree, and neural network. The models are evaluated on two selected roadway segments in St. Louis, MO, USA. Also, Bae et al. [10] developed a multi-contextual machine learning method to model the traffic impact of urban highway work zones. By adopting machine learning based approaches, these models can handle more complex work zone conditions compared to the simulation-based models. However, the performance is still constrained due to overly simplified model assumptions and model structures. These models either provide only aggregated traffic indicator prediction or focus narrowly on a specific aspect of traffic impact caused by work zone events.

Reviewing existing research highlights two major limitations in predicting the impact of work zone traffic. 1) The quality and quantity of data sources are limited, as there is often no comprehensive pipeline for integrating, curating, and

This work is supported by USDOT Intelligent Transportation Systems Joint Program Office under Grant 693JJ322F00355N.

Qinhua Jiang, Xishun Liao, Yaofa Gong, and Jiaqi Ma are with the Department of Civil and Environmental Engineering, University of California, Los Angeles, Los Angeles, USA.

E-mail: qhjiang93@ucla.edu; xishunliao@g.ucla.edu; gongyaofa0211@g.ucla.edu; jiaqima@ucla.edu;
*Corresponding author: xishunliao@g.ucla.edu

augmenting work zone traffic data for enhanced data-driven methods; 2) The data-driven methods currently used are overly simplified and not capable of handling the complex and dynamic traffic variations associated with work zones. Consequently, there is a pressing need for a model that can effectively capture the dependencies between spatial-temporal traffic patterns and work zone characteristics, providing a holistic perspective on both mobility and safety impacts.

The Work Zone Data Exchange (WZDx) Feed Registry, maintained by the U.S. Department of Transportation (USDOT), contains up-to-date metadata on work zone feeds that adhere to WZDx specifications [11]. Launched in 2019 by the Federal Highway Administration (FHWA) and the Intelligent Transportation Systems Joint Program Office (ITS JPO), this initiative seeks to enhance road safety and mobility by standardizing work zone data and ensuring its broad accessibility in a consistent format [12]. In this study, we utilize WZDx datasets from the ITS DataHub combined with the University of Maryland CATT Laboratory's Regional Integrated Transportation Information System (RITIS) data [12], providing insights into travel times and traffic speeds across Maryland's transportation network. Additionally, we integrate the Maryland Department of Transportation's (MDOT) Annual Average Daily Traffic statistics and loop detector data with incident data from RITIS and MDOT to create an enriched work zone dataset for predictive model training.

Besides the integration of multi-context datasets, the selection of data-driven models is crucial for estimating the traffic impact of work zones. Generally, traffic prediction models are categorized into short-term and long-term traffic forecasts [20][21][22][29]. Both use a sequence-to-sequence or sequence-to-one approach, where a sequence of past traffic readings from the previous $N$ timesteps is used to predict the traffic status for the following one or several timesteps, ranging from several minutes to multiple hours. These methods depend on the most recent traffic data to forecast future traffic conditions. However, these sequence-based models do not align with the objectives of our study. Our research aims to predict the traffic conditions on road segments with planned work zones well in advance before their implementation (e.g., days or weeks ahead), meaning no real-time traffic data at the time of making the prediction. To the best of the authors' knowledge, none of the existing traffic prediction models are designed for such goals. Inspired by image-based sequence-to-sequence traffic prediction methods [23][24][31], which transform city-level space-time traffic states into 2D images for model inputs and outputs, this paper proposes a novel image-to-image prediction method for work zone traffic forecast. This method converts the historical spatial-temporal traffic patterns into multi-channel image inputs and conducts a joint representation with the planned work zone features to deliver a comprehensive traffic prediction for the entire duration of the work zones at once, which enables the estimation of the traffic impact with high time resolution for the planning of work zones.

In summary, based on the curated dataset created by the data integration pipeline, we introduce an attention-based multi-context convolutional encoder-decoder neural network, named AMCNN-ED, to predict the impact of planned work zones, specifically focusing on mobility impact such as traffic speed, queue length, congestion start time/duration, and safety impact such as incident likelihood. The contributions of this paper are outlined as follows:

- Developed a data curation pipeline that integrates work zone event data with traffic and roadway network datasets, creating an enhanced data source tailored for predicting the traffic impacts of work zone events.

- Introduced an image-based modeling approach to estimate traffic impact caused by work zones by converting historical space-time traffic patterns into 2D images as model inputs. Based on that, we developed a novel attention-enhanced multi-context convolutional encoder-decoder neural network structure to capture the spatial-temporal dependencies between work zone characteristics and dynamic traffic patterns, enabling in-advance prediction of traffic impact (i.e., speed, queue length, congestion start time/duration, and incident likelihood) for planned work zones well ahead of time.

- Conducted a comprehensive evaluation of the proposed model using a real-world dataset from Maryland's transportation network, benchmarking it against baseline models to demonstrate its superior performance.

II. DATA INTEGRATION AND RECONSTRUCTION

*A. Multi-Context Work Zone Data*

To construct a comprehensive work zone dataset, this study follows the process depicted in Fig. 1, which outlines the integration of diverse datasets. WZDx provides dynamic and detailed work zone data, which allows for the extraction of precise work zone information, such as locations, timings, and specific characteristics like lane counts and geometries, as well as potential vehicular impacts. Traffic data from RITIS delivers link-level information, enriching the dataset with metrics like travel time and traffic speed, including historical and reference speeds. When combined with MDOT's volume data and supplemented by incident data, the dataset achieves a high level of granularity, covering individual work zones and their wider impact on the transportation network.

By incorporating these diverse data streams, we can acquire valuable operational metrics at the agranular level, encompassing individual work zones, their immediate surroundings, the impacted corridor, and the broader regional road network. Furthermore, the robust data capabilities will also facilitate more in-depth categorization based on different types of work zones and specific geographical regions. This enhanced categorization will provide us with an understanding

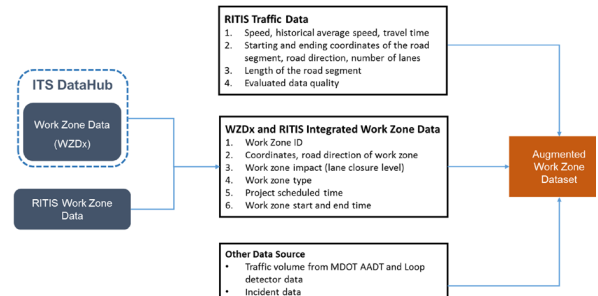

Figure 1. Multi-context data integration pipeline

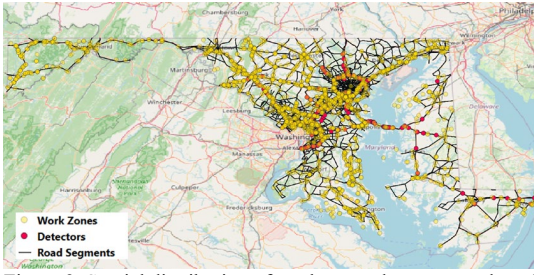

Figure 2. Spatial distribution of work zone, detectors, and road segments in Maryland transportation network

of the diverse impacts and dynamics across various work zone scenarios and geographic contexts.

*B. Data Integration and Space-Time Traffic Image Generation*

The integration of these datasets employs a sophisticated spatial-temporal matching process. As illustrated in Fig. 2, this map highlights the geospatial alignment of work zones, loop detectors, and road segments throughout the Maryland transportation network. Initially, matching is conducted using precise GPS coordinates to ensure each work zone is accurately paired with its corresponding road segment. Subsequently, traffic data of road segments and loop detectors relevant to the operational hours of each work zone are extracted. This dual-layered matching strategy — first spatial, then temporal — ensures a seamless amalgamation of location and time-specific traffic patterns.

To be specific, after cleaning and filtering, a total of 3646 work zones were identified from 2016 to 2019, excluding 2020 to 2022 due to the biased traffic patterns during the COVID-19 pandemic. The study focuses on temporary work zones with durations of less than 24 hours. To capture pertinent data for temporal work zone study and enable the development of an AI model, our approach is to compile data on a case-by-case basis for each work zone. For every individual case, we systematically collect traffic data encompassing the complete duration of the work zone event. In terms of spatial information, we include data for road segments extending 5 miles upstream relative to each work zone. Based on this spatial matching result, a feature of "distance to work zone" is calculated for each road segment; similarly, "time to work zone start" and "time after work zone end" are calculated for each time step. To ensure a high level of data fidelity, we maintain a time resolution of 15 minutes throughout the dataset.

As a result, for each work zone case, as shown in Fig. 3 (a), a 2D space-time matrix containing spatial-temporal information is organized, with the highlighted area indicating affected traffic. This matrix can be used to further represent other traffic features such as speed, historical average speed, and historical average volume, as well as geospatial features like link length and distance to work zone link, all updated in 15-minute intervals. These 2D space-time matrices are further converted to 2D heatmap images with different colors indicating different levels of values for particular traffic features, as seen in Fig. 3 (b). This systematic organization not only captures the real-time dynamics of work zones but also furnishes a standardized dataset format from which AI models can learn the complex spatial-temporal dependencies of traffic flow in relation to work zone activities, enhancing predictive capabilities.

### III. METHODOLOGY

*A. Problem Definition*

The methodology proposed by this paper tackles the problem of spatial-temporal traffic speed and incident likelihood prediction on road segments of planned work zone events well in advance before their implementation (e.g., days or weeks ahead). The definition of this predictive problem is presented as follows:

For a planned work zone event scheduled to start at $T_0$ and end at $T_n$ at location $L$ of a roadway, we define all the link segments on the same roadway within 5 miles upstream of $L$ as target links. Assume that for these target links, the historical average traffic sequences (e.g., traffic speed, volume) during the same time of day and day of week corresponding to the planned work zone schedules are known. Additionally, the geospatial correlations between the links (e.g., link length, link order) and the characteristics of the planned work zone event (e.g., number of closed lanes, number of total lanes, etc.) are also known. The model aims to predict two key outcomes: 1) a sequence of traffic speeds on all the target links throughout the duration of the planned work zone; and 2) the likelihood of an incident occurring on the target links during the work zone period. The spatial-temporal traffic speed output can further be used to infer other traffic impact attributes such as maximum queue length, congestion start time, and congestion duration.

*B. Model Structure*

(1) Model Overview

As shown in Fig. 4, the model input encompasses two components, including a set of historical traffic pattern and geospatial sequences which have been converted into a multi-channel 2D space-time image, and a tabular feature vector of work zone characteristics. The 2D space-time image consists mapping function $f: X \rightarrow Y$ that can predict the traffic speed for each timestep on each upstream link and the likelihood of incident occurrence during the work zone event.

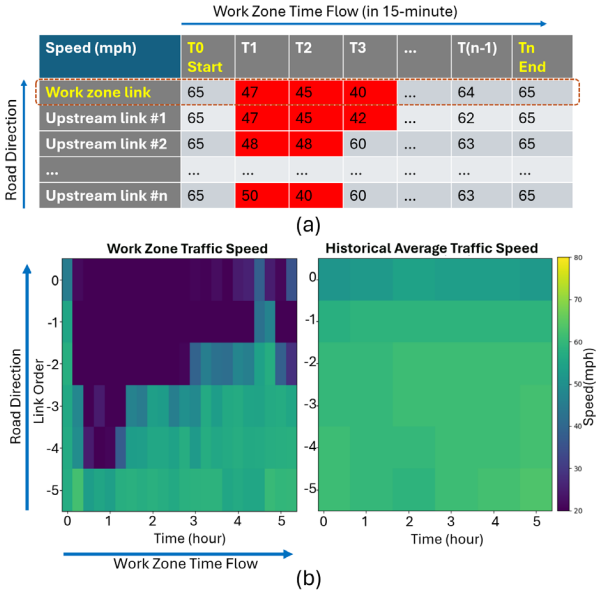

Figure 3. Work zone space-time traffic image generation: (a) 2D space-time traffic matrix; (b) Two examples of converted 2D space-time traffic images (work zone traffic speed and historical traffic speed)

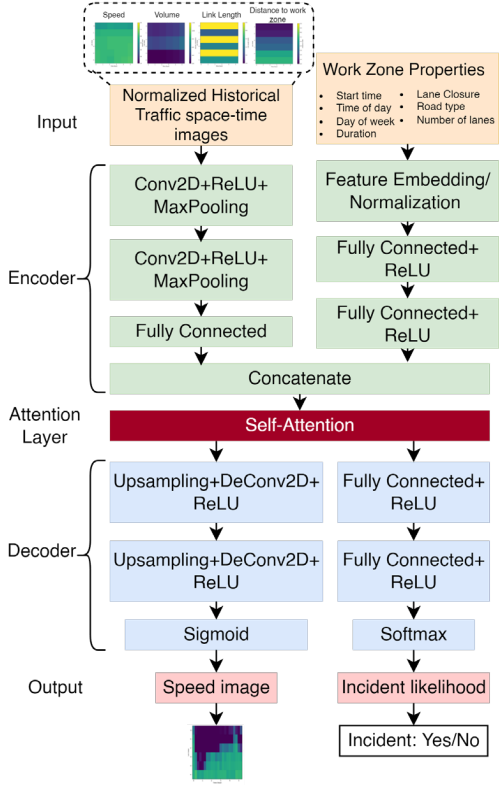

Figure 4. Model structure of AMCNN-ED

To model the work zone impact prediction problem, the AMCNN-ED model constructs 3 modules: encoder layers, attention layer, and decoder layers. The multi-context encoder extracts the spatiotemporal features from historical space-time traffic data and static work zone features from planned work zone tabular data. The extracted feature maps are combined and passed to the attention layer to weigh the importance of each part in the concatenated feature representation. Then the attention-enhanced feature vector is further sent to decoder layers with multiple transposed CNN layer and split in the output layer to generate both the 2D speed image and incident likelihood.

(2) Encoder Layers

The encoder consists of two parallel modules designed to create a joint representation of historical traffic information, geospatial features, and planned work zone characteristics. As shown in Fig. 4, the image encoder module employs two convolutional neural network (CNN) layers to extract spatial-temporal dependencies in the historical traffic patterns of upstream links during the work zone period. Each convolutional layer comprises a 2-dimensional convolution layer (Conv2D), a ReLU activation layer, and a max pooling layer, which collectively extract spatial-temporal features from the preceding layer. At the end of the two CNN layers, a flattened layer converts the feature map into a 1D vector representation. Additionally, a tabular feature extraction module extracts features from work zone-related attributes and converts them into a 1D feature vector, which can then be concatenated with the feature vector extracted from the CNN layer.

of multiple channels, each representing the historical traffic pattern and geospatial correlations of the link segments within the 5-mile range upstream of the work zone. The input image, $X_{image}$ can be defined as:

$$X_{image} \in \mathbb{R}^{h \times w \times c} = \{I_1^{h \times w}, I_2^{h \times w}, \ldots, I_c^{h \times w}\} \quad (1)$$

where $h$ refers to the height of the image, or the number of links within the work zone 5 miles range; $w$ refers to the width of the image, or the number of timesteps of the work zone event; $c$ is the number of channels of the input image, which is the number of features related to historical traffic pattern and geospatial relationships. In this study, we selected historical average speed, historical average volume, link length, and distance to work zone location as the four channels of the input image. The second input component is the feature vector of planned work zone characteristics, denoted as

$$X_{wz} \in \mathbb{R}^{n \times 1} = \{x_1, x_2, \ldots, x_n\} \quad (2)$$

where $n$ denotes the number of features of the work zone. In this study, we consider the following features: start time of day, day of week, work zone duration, number of lanes closed, number of total lanes, road type, and on-ramp/off-ramp connection.

$$Y = \{Y_{speed}, Y_{inci}\} \quad (3)$$

The output of the model, as denoted in (3), includes a predicted 2D space-time traffic speed image $Y_{speed} \in R^{h \times w \times 1}$ to indicate the speed of target links within 5 miles upstream of the anticipated work zone at 15-minute intervals for the work zone duration, and a likelihood that indicates the probability of incident occurrence during the projected work zone event. The goal of the work zone traffic impact prediction is to learn a

(3) Attention Layer

As presented in Fig. 4, the proposed network uses the self-attention mechanism to weigh the importance of different parts of the feature representation from the encoder layer. The self-attention mechanism is a deep learning technique originally designed for natural language processing (NLP) tasks to improve the modeling of relationships in sequential data [18], and further implemented in other areas such as helping the model to learn which part of the feature representation is more informative for succeeding model components [19][28].

When image and tabular data features are concatenated, they form a combined feature space. This space includes both the spatial information from the images and structured information from the tabular data. However, not all features contribute equally to the task at hand. An attention mechanism is employed here to dynamically learn to focus more on those features that are more relevant, effectively learning a task-specific weighting of features. By applying attention to the concatenated features, the model can highlight aspects of the data that are more informative for the specific prediction or reconstruction task. This selective focus can improve accuracy and robustness by reducing the impact of less relevant or noisy data.

To compute the decoder input, First, features extracted from both the image and the tabular data are combined into a single feature vector. This combined feature vector is then transformed into three different sets of vectors [19]: queries (**Q**), keys (**K**), and values (**V**). These transformations are achieved through multiplication by three distinct sets of weights. The model computes scores by comparing all the queries with all the keys. These scores determine how much attention or importance should be given to each value vector. Each value vector is then multiplied by its corresponding attention score, effectively emphasizing more important features and diminishing less important ones. The resulting weighted sum forms a new, attention-enhanced feature vector that is used as the input for the decoder. The process of implementing the self-attention mechanism on the encoded input feature can be expressed by the following equations:

$$\begin{cases} [\mathbf{Q}, \mathbf{K}, \mathbf{V}] = [\mathbf{W_Q}, \mathbf{W_K}, \mathbf{W_V}] \cdot x \\ \mathbf{A} = \text{softmax}\left(\frac{\mathbf{Q} \cdot \mathbf{K}^T}{\sqrt{d_k}}\right) \\ \mathbf{z} = \mathbf{AV} \end{cases} \quad (4)$$

where $x$ is the concatenated input feature vector from the encoder layer, $\mathbf{W_Q}, \mathbf{W_K}$, and $\mathbf{W_V}$ are weight matrices, $\sqrt{d_k}$ is a scaling factor, and $\mathbf{z}$ is the output feature vector after applying self-attention.

(4) Decoder Layers

The decoder layer of this network consists of two transposed convolution layers for image reconstruction and a set of fully connected layers for incident likelihood prediction. The transposed convolution layers, denoted as DeCNN, are used to reconstruct the encoded feature vector to produce a 2D image for speed prediction.

The decoder receives an attention-enhanced feature vector from the attention layer. Then, the first DeCNN layer takes the flattened feature vector from the previous layer and reshapes it back into a multi-dimensional tensor. It then applies transposed convolution operations to start upsampling the features back to the spatial dimensions needed for image reconstruction. Following the initial upsampling, the second DeCNN layer further increases the spatial dimensions of the feature map, continuing to add detail and structure. It reduces the number of channels, aiming to reconstruct the spatial structure of the original input image. After each transposed convolution, an activation function such as ReLU is applied to introduce non-linearity, helping to model complex patterns in the data.

Following two consecutive DeCNN layers, the image output path employs a sigmoid activation to normalize the image pixels for the one-channel speed graph. On the classification output side, the attention-enhanced feature vector is sent to a set of fully connected layers with a softmax activation function at the end to output a probability between 0 and 1, indicating the likelihood of the input belonging to one of two incident labels.

*C. Loss Function*

Given that the model adopts a multi-tasking learning structure and outputs two types of outputs, i.e., 2D space-time traffic speed image and incident likelihood, we employ distinct loss functions for each target output and combine them to represent the model's overall loss.

For the traffic speed prediction, we implement the widely used Huber loss function to mitigate the impact of outliers in speed predictions [25]. The definition of Huber loss is provided in (5), where $y$ and $\hat{y}$ refer to observed and predicted speeds, respectively, and $\delta$ is a hyperparameter that requires tuning.

$$L_\delta(y, \hat{y}) = \begin{cases} \frac{1}{2}(y - \hat{y})^2 & \text{for } |y - \hat{y}| \leq \delta, \\ \delta|y - \hat{y}| - \frac{1}{2}\delta^2 & \text{otherwise.} \end{cases} \quad (5)$$

For incident prediction, we employ cross-entropy loss, commonly used in classification problems [10], denoted as $L_{ce}$. This loss function measures the performance of the classification output, which is a probability value between 0 and 1. The total loss can be expressed as the weighted sum of the losses from the two tasks:

$$L = w_1 \cdot L_\delta + w_2 \cdot L_{ce} \quad (6)$$

where weights $w_1$ and $w_2$ are hyperparameters to be tuned during model training.

IV. EXPERIMENTATION

*A. Performance Metrics*

The experimentation adopts three widely applied evaluation metrics to quantify the performance of speed prediction of each model [18][17][32]. They are Mean Square Error (MSE), Root Mean Square Error (RMSE), and Mean Absolute Percentage Error (MAPE). The performance metrics are presented in (5), where $\hat{y}_i$ represents the predicted speed made by the model, $y_i$ represents the corresponding ground-truth value.

$$\begin{cases} RMSE = \sqrt{\frac{1}{N}\sum_{i=1}^{N}(\hat{y}_i - y_i)^2} \\ MAE = \frac{1}{N}\sum_{i=1}^{N}|\hat{y}_i - y_i| \\ MAPE = \frac{1}{N}\sum_{i=1}^{N}\frac{|\hat{y}_i - y_i|}{y_i} \end{cases} \quad (7)$$

Regarding incident prediction, we adopted three commonly used classification prediction metrics, recall, precision, and F1 score to assess each model's performance [30]. Recall measures the proportion of actual positives correctly identified by the model, highlighting its sensitivity. Precision assesses the accuracy of the positive predictions made by the model, indicating the proportion of true positives among all positive predictions. F1 score is the harmonic mean of precision and recall, providing a single metric that balances both the precision and the recall to measure a model's accuracy more comprehensively.

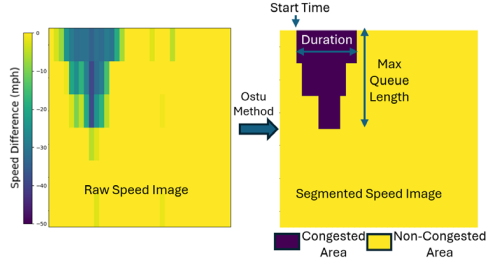

Figure 5. Work zone space-time traffic image processing

For work zone samples exhibiting congestion patterns, we introduced three congestion-specific metrics to evaluate the prediction performance: the start time, duration, and maximum queue length of the congestion, where the duration and queue length are the width and depth of the congestion area on the space-time speed image. For each 2D space-time image corresponding to a work zone, to minimize the interference of random data noise, we only consider congestion that last over one hour and extend across multiple consecutive link segments as valid. To identify valid congestion areas in the space-time images, we employed Otsu's method, an automated process used widely in image segmentation. Otsu's thresholding algorithm, a popular technique in image processing, is particularly effective for automatically performing clustering-based image thresholding [13]. The method operates by calculating the histogram of the pixel intensities and systematically testing all possible thresholds to determine which maximizes the between-class variance (i.e., the variance between the pixel intensities above and below the threshold) [14].

B. Baseline Models

The results of our model are compared against the following models:

- ARIMA: Auto-regressive integrated moving average.
- GRU: Gated recurrent unit network
- LSTM: Long-short-term memory network
- Conv-LSTM: Convolutional long-short-term memory network

The first four models—ARIMA [26], GRU [27], LSTM [27], and Conv-LSTM [23]—all make speed predictions in an autoregressive form. They require a short initial sequence as input to predict the very first timestep during the work zone. They then gradually append the newly predicted speed values to the input sequence and use the extended sequence to predict the next timestep until the entire duration of the work zone is predicted. The MCNN-ED model uses the same encoder-decoder structure as the AMCNN-ED proposed by this study, the only difference being that MCNN-ED does not incorporate a self-attention layer to enhance the feature representation. It should be noted that there aren't any existing models that can be applied directly for the problem defined in this study, therefore the baseline models listed here are highly customized to fit the work zone prediction scenario in this paper, the literature cited here only provided high-level concepts instead of complete model structures.

All neural network models were implemented using Pytorch 2.0. Each model was trained on a RTX A5000 GPU, providing ample GPU memory to facilitate the learning process. Additionally, the Adam optimizer was employed. The models were run for 200 epochs. Early stopping was implemented to prevent overfitting, halting the training process if the validation loss deteriorated for a specified number of epochs, even if the training loss continued to decrease.

TABLE I. SPEED PREDICTION RESULTS

| Model | | Performance metrics | | |
|---|---|---|---|---|
| | | *MAE* | *RMSE* | *MAPE* |
| Auto-Regression | ARIMA | 10.83 | 12.11 | 17.72 |
| | GRU | 8.14 | 8.63 | 14.65 |
| | LSTM | 7.99 | 8.61 | 14.53 |
| | Conv-LSTM | 7.59 | 8.46 | 13.83 |
| Encoder-Decoder | MCNN-ED | 7.36 | 8.30 | 13.77 |
| | **AMCNN-ED** | **7.10** | **8.06** | **13.16** |

V. RESULTS AND PERFORMANCE EVALUATION

A. System-Level Performance Analysis

Table I, II, and III present the prediction results of the proposed model and baseline models on the testing dataset for the 547 work zone events. Table I displays the results on all test work zones, while Table II focuses on results on congested areas of impacted work zones. The results in Table I demonstrate that the neural network-based models all outperform the ARIMA model. This is because ARIMA relies solely on previous timesteps' traffic data and fails to account for changes in traffic caused by work zone activities. Additionally, the results indicate that RNN-based autoregressive models do not perform as well as encoder-decoder structures, due to their inability to capture the comprehensive spatial-temporal dependencies between work zone properties and traffic patterns. Among the two encoder-decoder models, AMCNN-ED outperforms CNN-ED. This superior performance can be attributed to the self-attention layer in AMCNN-ED, which enhances the model's ability to discern the relative importance of different sectors in the joint feature representations produced by the encoder layers.

From the perspective of in-advance traffic management and long-term work zone planning, accurately forecasting the road segments impacted by work zone activities is of paramount importance. Therefore, we selected 50 work zone samples that experienced congestion during the work zone duration from a total of 547 test work zones to compare the performance between our model and baseline models. The ARIMA model, unable to predict traffic congestion caused by work zone activities, was excluded from the analysis in Table II.

As shown in Table II, compared to RNN-based autoregressive models, the two encoder-decoder approaches demonstrate substantial improvements. This suggests that multi-context convolutional feature extraction is more effective at capturing the spatial-temporal correlations across multiple adjacent locations over extended periods. This capability is particularly crucial for predicting non-recurrent congestion patterns during work zone events. Furthermore, the AMCNN-ED structure outperforms the CNN-ED structure, primarily due to its self-attention mechanism, which enables the model to identify key elements in the feature vectors from both the static work zone features and the historical spatial-temporal traffic patterns, thus more accurately predicting the occurrence of traffic congestion.

TABLE II. CONGESTED AREA PREDICTION RESULTS

| Model | | RMSE | | |
|---|---|---|---|---|
| | | Congestion Start Time | Congestion Duration | Max Queue Length |
| Auto-Regression | GRU | 2.32 | 2.97 | 1.53 |
| | LSTM | 2.24 | 2.99 | 1.45 |
| | ConvLSTM | 2.18 | 2.88 | 1.36 |
| Encoder-Decoder | MCNN-ED | 2.03 | 2.65 | 1.22 |
| | **AMCNN-ED** | **1.92** | **2.53** | **1.09** |

Table III presents the prediction results for collision incidents during work zone events. We excluded ARIMA from the model list since it is designed solely for time-series prediction and cannot provide classification outputs. The results show that the AMCNN-ED model outperforms the baseline model across all three performance metrics. This suggests that the AMCNN-ED model is more effective at predicting potential collision incidents compared to autoregressive models and non-attention-based encoder-decoder models, while also minimizing false alarms in work zones. It should be noted that the prediction accuracy of all the listed models remains below 0.7. This limitation is largely due to the stochastic nature of incidents and the current limitations of available data. According to various studies [15][16][30], the occurrence of collisions is influenced by numerous factors, including traffic, road closures, and external conditions such as weather, driver behavior, and vehicle conditions. Therefore, it is challenging to achieve precise forecasts for incident occurrences based solely on historical traffic data and projected work zone properties. However, the results demonstrated by this model still show promising potential to assist in the prevention of potential crashes during the planning of work zone activities.

### B. Event-Level Spatial-Temporal Performance Analysis

In this section, we focus on the model performance of selected examples from test dataset to illustrate the prediction performance of the proposed AMCNN-ED model and compare it with the best-performing baseline model at the event level in Fig. 6. Each plot in Fig. 6 represents a 2D space-time speed difference graph for the duration of each work zone. In Fig. 6, the ground truth speed graph is displayed in the left column, the prediction results from AMCNN-ED are in the middle column, and the results from the best-performing autoregressive model (Conv-LSTM) are in the right column. The dark blue areas indicate significant speed drops compared to the historical average speed at the same time of day, signaling severe congestion, while the yellow areas indicate speeds similar to the historical average. As shown in Fig. 6, the four work zones caused one or more instances of congestion during the work zone period, extending to multiple link segments upstream. The AMCNN-ED model more accurately captures the timing and spatial extent of the congestion compared to the autoregressive model. In contrast, the Conv-LSTM model tends to underpredict or overpredict the congestion area. A key insight from this comparison is that autoregressive models may incorrectly interpret traffic flow's temporal variations. This occurs because they predict each timestep based solely on previous timesteps, ignoring shockwave propagation. In contrast, the AMCNN-ED model incorporates global information, both temporally and spatially, for the work zone event. This underscores the benefits of using an attention-based encoder-decoder structure over an autoregressive structure for predicting traffic patterns during planned work zone activities.

TABLE III. INCIDENT PREDICTION RESULTS

| Model | | Performance metrics | | |
|---|---|---|---|---|
| | | Recall | Precision | F1 Score |
| Auto-Regression | GRU | 0.54 | 0.62 | 0.58 |
| | LSTM | 0.53 | 0.61 | 0.57 |
| | ConvLSTM | 0.54 | 0.62 | 0.58 |
| Encoder-Decoder | MCNN-ED | 0.56 | 0.62 | 0.58 |
| | **AMCNN-ED** | **0.58** | **0.65** | **0.61** |

### VI. CONCLUSION

This paper presents a data curation pipeline for data-centric work zone traffic prediction problems and proposes an attention-based multi-context encoder-decoder convolutional model to predict the traffic impact of planned work zone events. Our method consists of two main steps. First, we integrated archived data from multiple data platforms to construct a curated work zone traffic dataset that encompasses the essential factors influencing traffic changes and work zone characteristics. Next, we developed a convolutional encoder-decoder model to create a joint representation of multi-context spatial-temporal input features and implemented a self-attention mechanism to highlight key sectors within the encoded features. These features are then reconstructed through the transposed convolutional decoder layers to generate predictions for traffic speed and incident likelihood during the work zone events. The model, evaluated using four years of archived traffic data from Maryland, reduces the prediction error of traffic speed by 5% to 34%, queue length by 11% to 29%, congestion timing by 6% to 17%, and increases the accuracy of incident predictions by 5% to 7% compared to baseline models. Future research could extend the algorithm's application to other regions and assess its potential to enhance prediction performance for other non-recurrent traffic conditions.

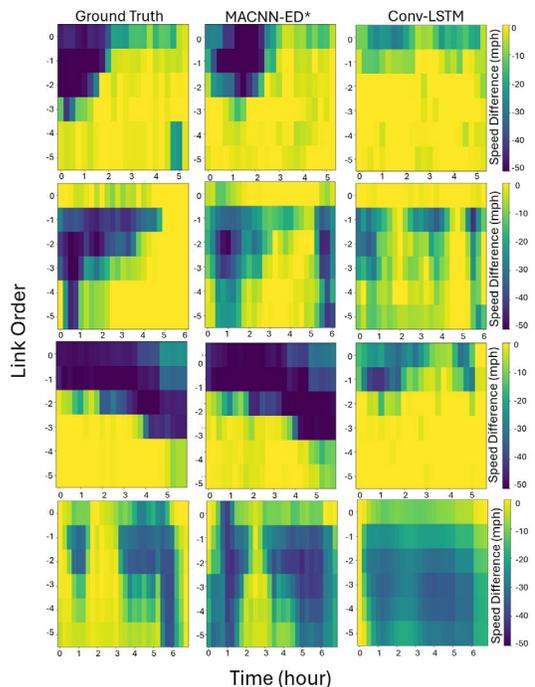

Figure 6. Examples of event-level speed prediction performance

ACKNOWLEDGMENT

We would like to thank Center of Excellence on New Mobility and Automated Vehicles (Mobility COE) for its support on this study.

REFERENCES

[1] "Making Work Zones Work Better," Federal Highway Administration (FHWA), 2004. [Online]. Available: https://ops.fhwa.dot.gov/aboutus/one_pagers/wz.htm
[2] "FHWA Work Zone Facts and Statistics," Federal Highway Administration (FHWA), 2024. [Online]. Available: https://ops.fhwa.dot.gov/wz/resources/facts_stats.htm
[3] "Using Modeling and Simulation Tools for Work Zone Analysis," Federal Highway Administration (FHWA), 2009. [Online]. Available: https://ops.fhwa.dot.gov/wz/traffic_analysis/wza_leaflet/
[4] W. v Ping and K. Zhu, "Evaluation of work zone capacity estimation models: A computer simulation study," In *Sixth Asia-Pacific Transportation Development Conference, 19th ICTPA Annual Meeting*. 2006.
[5] P. Chatterjee, P. Edara, S. Menneni, and C. Sun, "Replication of work zone capacity values in a simulation model," *Transportation Research Record*, vol. 2130, no. 1, pp. 138–148, 2009.
[6] X. Wen, "A work zone simulation model for travel time prediction in a connected vehicle environment," arXiv preprint arXiv:1801.07579, 2018.
[7] H. Adeli and X. Jiang, "Neuro-fuzzy logic model for freeway work zone capacity estimation," *Journal of Transportation Engineering*, vol. 129, no. 5, pp. 484–493, 2003.
[8] Karim and H. Adeli, "Radial basis function neural network for work zone capacity and queue estimation," *Journal of Transportation Engineering*, vol. 129, no. 5, pp. 494–503, 2003.
[9] Y. Hou, P. Edara, and C. Sun, "Traffic flow forecasting for urban work zones," *IEEE Transactions on Intelligent Transportation Systems*, vol. 16, no. 4, pp. 1761–1770, 2014.
[10] Bae, J., K. Choi, and J. H. Oh, "Multicontextual Machine Learning Approach to Modeling Traffic Impact of Urban Highway Work Zones," *Transportation Research Record: Journal of the Transportation Research Board*, 2017, 2645: 184–194.
[11] "Work Zone Data Exchange (WZDx)," Federal Highway Administration (FHWA), 2022. [Online]. Available: https://ops.fhwa.dot.gov/wz/wzdx/index.htm
[12] "RITIS: The Regional Integrated Transportation Information System," The Center for Advanced Transportation Technology (CATT) Lab, 2024. [Online]. Available: https://www.cattlab.umd.edu/ritis/
[13] S. L. Bangare, A. Dubal, P. S. Bangare, and S. Patil, "Reviewing Otsu's method for image thresholding," *International Journal of Applied Engineering Research*, vol. 10, no. 9, pp. 21777-21783, 2015.
[14] X. Liao, G. Wu, L. Yang, and M. J. Barth, "A real-world data-driven approach for estimating environmental impacts of traffic accidents," *Transportation Research Part D: Transport and Environment*, vol. 117, p. 103664, 2023.
[15] Q. Meng and J. Weng, "Evaluation of rear-end crash risk at work zone using work zone traffic data," *Accident Analysis & Prevention*, vol. 43, no. 4, pp. 1291-1300, 2011.
[16] S. Mokhtarimousavi, J. C. Anderson, A. Azizinamini, and M. Hadi, "Improved support vector machine models for work zone crash injury severity prediction and analysis," *Transportation Research Record*, vol. 2673, no. 11, pp. 680-692, 2019.
[17] Q. Jiang, B. Schroeder, J. Ma, L. Rodegerdts, B. Cesme, A. Bibeka, and A. Morgan, "Developing Highway Capacity Manual capacity adjustment factors for connected and automated traffic on roundabouts," *Journal of Transportation Engineering, Part A: Systems*, vol. 148, no. 5, 04022014, 2022.
[18] A. Abdelraouf, M. Abdel-Aty, and J. Yuan, "Utilizing attention-based multi-encoder-decoder neural networks for freeway traffic speed prediction," *IEEE Transactions on Intelligent Transportation Systems*, vol. 23, no. 8, pp. 11960-11969, 2021.
[19] S. Zhang, C. Zhang, S. Zhang and J. J. Q. Yu, "Attention-Driven Recurrent Imputation for Traffic Speed," in *IEEE Open Journal of Intelligent Transportation Systems*, vol. 3, pp. 723-737, 2022
[20] X. Yin, G. Wu, J. Wei, Y. Shen, H. Qi, and B. Yin, "Deep learning on traffic prediction: Methods, analysis, and future directions," *IEEE Transactions on Intelligent Transportation Systems*, vol. 23, no. 6, pp. 4927-4943, 2021.
[21] M. Akhtar and S. Moridpour, "A review of traffic congestion prediction using artificial intelligence," *Journal of Advanced Transportation*, 2021, pp. 1-18.
[22] M. Zhong, J. Kim and Z. Zheng, "Estimating Link Flows in Road Networks With Synthetic Trajectory Data Generation: Inverse Reinforcement Learning Approach," in *IEEE Open Journal of Intelligent Transportation Systems*, vol. 4, pp. 14-29, 2023
[23] N. Ranjan, S. Bhandari, H. P. Zhao, H. Kim, and P. Khan, "City-wide traffic congestion prediction based on CNN, LSTM and transpose CNN," *IEEE Access*, vol. 8, pp. 81606-81620, 2020.
[24] D. Jo, B. Yu, H. Jeon, and K. Sohn, "Image-to-image learning to predict traffic speeds by considering area-wide spatio-temporal dependencies," *IEEE Transactions on Vehicular Technology*, vol. 68, no. 2, pp. 1188-1197, 2018.
[25] M. Tang, X. Fu, H. Wu, Q. Huang, and Q. Zhao, "Traffic flow anomaly detection based on robust ridge regression with particle swarm optimization algorithm," *Mathematical Problems in Engineering*, 2020, pp. 1-10.
[26] B. Dissanayake, O. Hemachandra, N. Lakshitha, D. Haputhanthri, and A. Wijayasiri, "A comparison of ARIMAX, VAR and LSTM on multivariate short-term traffic volume forecasting," in *Conference of Open Innovations Association*, FRUCT, no. 28, pp. 564-570, FRUCT Oy, 2021.
[27] R. Fu, Z. Zhang, and L. Li, "Using LSTM and GRU neural network methods for traffic flow prediction," in 2016 *31st Youth academic annual conference of Chinese association of automation (YAC)*, pp. 324-328, IEEE, Nov. 2016.
[28] Z. Wang, J. Guo, Z. Hu, H. Zhang, J. Zhang and J. Pu, "Lane Transformer: A High-Efficiency Trajectory Prediction Model," in *IEEE Open Journal of Intelligent Transportation Systems*, vol. 4, pp. 2-13, 2023,
[29] J. Desai, B. Scholer, J. K. Mathew, H. Li and D. M. Bullock, "Analysis of Route Choice During Planned and Unplanned Road Closures," in *IEEE Open Journal of Intelligent Transportation Systems*, vol. 3, pp. 489-502, 2022
[30] M. Emu, F. B. Kamal, S. Choudhury and Q. A. Rahman, "Fatality Prediction for Motor Vehicle Collisions: Mining Big Data Using Deep Learning and Ensemble Methods," in *IEEE Open Journal of Intelligent Transportation Systems*, vol. 3, pp. 199-209, 2022
[31] Papa, I. Cardei and M. Cardei, "Generalized Path Planning for UTM Systems With a Space-Time Graph," in *IEEE Open Journal of Intelligent Transportation Systems*, vol. 3, pp. 351-368, 2022
[32] Q. Jiang, D. Nian, Y. Guo, M. Ahmed, G. Yang, and J. Ma, "Evaluating connected vehicle-based weather responsive management strategies using weather-sensitive microscopic simulation," *Journal of Intelligent Transportation Systems*, vol. 27, no. 1, pp. 92-110, 2023.